\documentclass{article} 
\usepackage{iclr2026_conference,times}

\usepackage{amsmath,amsfonts,bm}

\def\eqref#1{equation~\ref{#1}}

\def\1{\bm{1}}

\DeclareMathAlphabet{\mathsfit}{\encodingdefault}{\sfdefault}{m}{sl}
\SetMathAlphabet{\mathsfit}{bold}{\encodingdefault}{\sfdefault}{bx}{n}

\newcommand{\KL}{D_{\mathrm{KL}}}

\usepackage{hyperref}
\usepackage{url}
\usepackage{subcaption}
\usepackage{cleveref}
\usepackage{amssymb}
\usepackage{svg}

\usepackage{tikz}
\usetikzlibrary{bayesnet,positioning,arrows.meta}

\usepackage{booktabs}
\usepackage{siunitx}
\usepackage{multirow}
\usepackage{booktabs,tabularx,makecell}

\title{Latent Action World Models for Control with Unlabeled Trajectories}

\author{Marvin Alles$^{1\thanks{Corresponding author.}}$ 
\quad Xingyuan Zhang$^{1, 2}$ 
\quad Patrick van der Smagt$^{3, 4}$
\quad Philip Becker-Ehmck$^{2}$ \\
\quad $^1$Technical University of Munich \quad
$^2$Machine Learning Research Lab, Volkswagen Group \\
\qquad \qquad \qquad $^3$Eötvös Loránd University Budapest \quad
$^3$Foundation Robotics Labs \\
}

\iclrfinalcopy 
\begin{document}

\maketitle

\begin{abstract}
Inspired by how humans combine direct interaction with action-free experience (e.g., videos), we study world models that learn from heterogeneous data. Standard world models typically rely on action-conditioned trajectories, which limits effectiveness when action labels are scarce. We introduce a family of latent-action world models that jointly use action-conditioned and action-free data by learning a shared latent action representation. This latent space aligns observed control signals with actions inferred from passive observations, enabling a single dynamics model to train on large-scale unlabeled trajectories while requiring only a small set of action-labeled ones. 
We use the latent-action world model to learn a latent-action policy through offline reinforcement learning (RL), thereby bridging two traditionally separate domains: offline RL, which typically relies on action-conditioned data, and action-free training, which is rarely used with subsequent RL.
On the DeepMind Control Suite, our approach achieves strong performance while using about an order of magnitude fewer action-labeled samples than purely action-conditioned baselines. These results show that latent actions enable training on both passive and interactive data, which makes world models learn more efficiently.
\end{abstract}

\section{Introduction}
Deep-learning has achieved remarkable success in computer vision and natural language processing, largely enabled by abundant datasets. In contrast, progress in real-world control, especially with deep reinforcement learning (RL), remains limited. The main reasons are sample inefficiency of RL and the need for costly and unsafe environment interaction \citep{ChallengesRealWorldRL, Kober2013ReinforcementLI}. Offline RL removes the need of interacting with the environment by learning from static datasets \citep{BatchRLErnst,BatchRLLange,OfflineRLTutorialReview}. Still, two challenges persist: (i) high data requirements, and (ii) the scarcity of robot datasets with action labels. At the same time, large amounts of unlabeled trajectories (e.g., videos) are readily available. Using such action-free data for RL is desirable, but usual RL methods typically require action annotations.

Humans do not learn only from interaction, but also from observation. Building on this premise, we argue that using action-free data requires a causal representation of actions and their effects, similar to human understanding. Such a representation can be learned with world models. However, typically, in order to learn the dynamics of the environment, they assume access to actions. We therefore hypothesize that an explicit latent action representation is crucial: it enables world models to use both action-conditioned and action-free experience. Concretely, we introduce a family of latent action world models (LAWM) that jointly train on mixed data sources, using a small set of action-labeled trajectories when available, and inferring latent actions via inverse dynamics when they are not. 
With action labels, our approach builds on established work on latent-actions from the offline RL literature \citep{c-lap, PLAS}; without labels, it infers actions from future observations to align the latent action space across both settings. The result is a unified latent action representation that serves as an effective inductive bias for downstream policy learning and reduces the reliance on large labeled datasets.

In the literature on world models, reinforcement learning, and latent actions, two main research streams have emerged. The first, focusing on Offline RL, introduces latent actions as an inductive bias: a generative model is trained to capture the dataset's action distribution and then used to constrain the learned policy to that distribution \citep{PLAS, c-lap, Chen2022LAPOLA, LetOR}. This line of work relies on action-conditioned RL datasets. The second learns from action-free videos, typically combining forward and inverse dynamic models to learn a latent action representation, which is then used for downstream behavior cloning \citep{schmidt2024learningactactions, ye2025latentactionpretrainingvideos, Gao2025AdaWorldLA, bruce2024geniegenerativeinteractiveenvironments, baker2022videopretrainingvptlearning, cui2024dynamoindomaindynamicspretraining}. To date, these streams largely proceed in parallel: action-free training is rarely paired with subsequent RL, while Offline RL work depends on action-conditioned data.

With LAWM, we aim to combine both research directions by introducing a family of latent-action world models that can be trained on both action-free and action-conditioned data, thereby unifying these paradigms. The resulting models learn latent action representations that are a strong inductive bias for downstream reinforcement learning.

\begin{figure}[t]
    \centering
    \includegraphics[width=0.6\columnwidth]{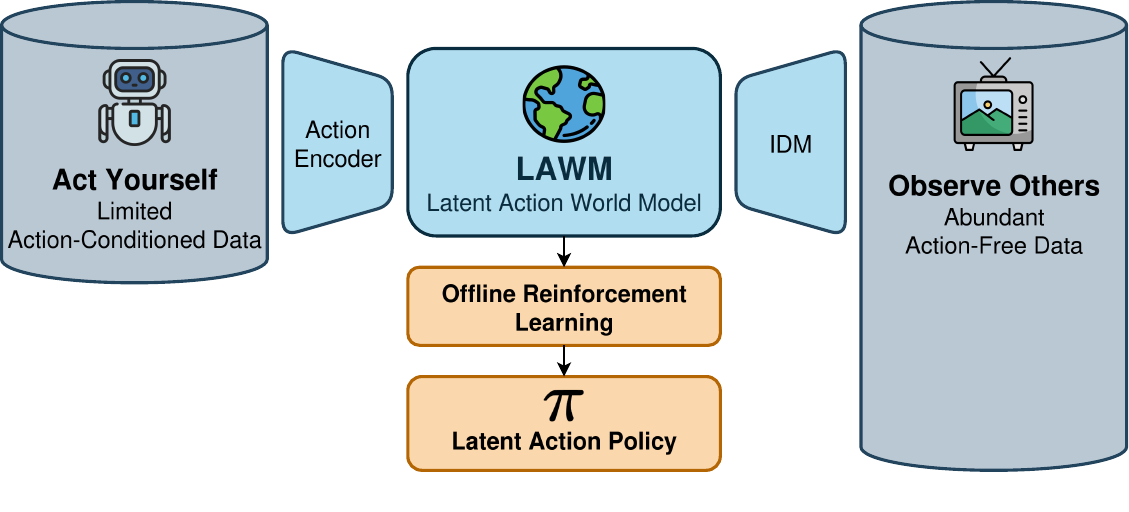}
    \caption{Overview of our setting and general approach. We assume the existence of a large action-free dataset and a small action-conditioned dataset. LAWM can combine both data sources using a latent action representation. The model can then be used for offline reinforcement learning, in particular as proposed by C-LAP \cite{c-lap}.}
    \label{fig:overview}
\end{figure}

\section{Problem Definition}
We consider an RL setting with static datasets (also referred to as offline RL or batch RL \citep{BatchRLErnst, BatchRLLange, OfflineRLTutorialReview}), but where datasets only contain a small fraction of action-labeled trajectories. Concretely, we define the action-conditioned dataset as $\mathcal{D}_{act} = \{(o_{1:T}, a_{1:T}), ... \}$ and the action-free dataset as $\mathcal{D}_{\varnothing} = \{(o_{1:T}), ...\}$. Note that, both datasets contain rewards, which we omit here for simplicity. The complete static dataset is thus defined as $\mathcal{D} = \mathcal{D}_{act} \cap \mathcal{D}_{\varnothing}$, with $| \mathcal{D}_{act} | \ll | \mathcal{D}_{\varnothing} |$.

\section{Latent Action World Models}

We introduce a family of world models that can learn from datasets which contain both action-free and action-conditioned trajectories. 
To achieve this, we use a latent action space to align action representations of both settings. 
The latent action representation serves both as an inductive bias, when learning from action-free data, and as a natural support constraint when the model is later used for planning or policy learning.
To learn a latent action world model, we consider two settings depending on whether the data contains action trajectories. If actions are available, we train an action-conditioned latent-action model (\Cref{fig:graphical_model_action_conditioned}); if not, we train an action-free variant (\Cref{fig:graphical_model_action_free}). Both formulations share the same prior; they differ only in the latent-action inference mechanism.

\paragraph{Action-conditioned latent action world model}
Given joint observation-action trajectories $\mathcal{D}_{act} = \{(o_{1:T}, a_{1:T}), ... \}$ as in offline reinforcement learning, we follow \cite{c-lap}, and define a generative model of observations and actions with the following Markovian assumptions.
\begin{align}
 p(o_{1:T}, a_{1:T})
 &= \int p(o_{1:T}, a_{1:T}, s_{1:T}, u_{1:T})~ds~du \\
 & = \int p(s_0) \prod_{t=1}^T p_{\theta}(o_t | s_t) p_{\theta}(a_t | s_t, u_t) p_{\theta}(u_{t}| s_{t}) p_{\theta}(s_{t} | s_{t-1}, a_{t-1}) ~ds~du
\label{eq:p_ac}
\end{align}
We use variational inference to learn the model, and therefore introduce 
\begin{align}
q_{\phi}(s_{1:T}, u_{1:T} \mid o_{1:T}, a_{1:T})
= q(s_0) \prod_{t=1}^T
q_{\phi}(s_t \mid s_{t-1}, a_{t-1}, o_t)\,
q_{\phi}(u_t \mid s_t, a_t)
\label{eq:q_ac}
\end{align}
as inference models alongside generative models $p_{\theta}$.
The following provides an overview of the components:
\begin{align}
\nonumber
\begin{aligned}
&\text{latent state prior} && p_{\theta}(s_{t} | s_{t-1}, a_{t-1})\\
&\text{latent state posterior} && q_{\phi}(s_{t} | s_{t-1}, a_{t-1}, o_{t})\\
&\text{observation decoder} && p_{\theta}(o_t | s_t)
\end{aligned}
\qquad\qquad
\begin{aligned}
&\text{latent action prior} && p_{\theta}(u_{t}| s_{t})\\
&\text{latent action posterior} && q_{\phi}(u_t | s_t, a_t) \\
&\text{action decoder} && p_{\theta}(a_t | s_t, u_t)
\end{aligned}
\end{align}

Compared to model-based methods, which learn a model of the conditional distribution $p(o_{1:T} | a_{1:T})$ \citep{hafner2019learninglatentdynamicsplanning}, the above factorization introduces latent actions  $u_t$ alongside latent states $s_t$. This formulation is useful in offline reinforcement learning: We can constrain the policy to the latent action prior to only generate actions within support of the dataset's action distribution. This is effective to address the common issue of value overestimation in offline reinforcement learning \citep{c-lap}.

\paragraph{Action-free latent action world model}
In many practical settings, action annotations are unavailable and only observation trajectories $\mathcal{D}_{\varnothing} = \{(o_{1:T}), ...\}$ are provided. To handle such cases, we can use the latent action factorization as inductive bias. We provide an action-free latent action world model that uses the same generative formulation, but modifies the inference model. We define the generative model as  
\begin{align}
 p(o_{1:T})
 &= \int p(o_{1:T}, a_{1:T}, s_{1:T}, u_{1:T})~da~ds~du \\
 & = \int p(s_0) \prod_{t=1}^T p_{\theta}(o_t | s_t) p_{\theta}(a_t | s_t, u_t) p_{\theta}(u_{t}| s_{t}) p_{\theta}(s_{t} | s_{t-1}, a_{t-1}) ~da~ds~du.
\label{eq:p_af}
\end{align}
Note that different to the action-conditioned generative model \Cref{eq:p_ac} we marginalize over actions $a_t$. To infer latent actions in the absence of actions in the dataset we use the following inference model:
\begin{align}
q_{\phi}(s_{1:T}, u_{1:T}, a_{1:T} \mid o_{1:T})
= q(s_0, a_0) \prod_{t=1}^T
q_{\phi}(s_t \mid s_{t-1}, a_{t-1}, o_t)\,
q_{\phi}(a_t | s_t, u_t)\,
q_{\phi}(u_t \mid s_t, o_{t+1}).
\label{eq:q_af}
\end{align}
Instead of using a latent action posterior of the form $q_{\phi}(u_t | s_t, a_t)$ as in the action-conditioned inference model \Cref{eq:q_ac}, we infer latent actions only from observations. Therefore, we use future observation $o_{t+1}$ as a weak supervisory signal of the action-free latent action posterior $q_{\phi}(u_t | s_t, o_{t+1})$. This can be interpreted as a form of inverse dynamics modeling \citep{baker2022videopretrainingvptlearning}. We provide further details on the training objective in \Cref{sec:training_objective}.

\section{Experiments}
We evaluate LAWM on the DeepMind Control Suite, focusing on the cheetah-run, walker-walk, and hopper-stand environments. For each environment we use four datasets namely medium, medium-replay, expert, plan2explore to analyse the influence of dataset properties such as narrowness of the distribution (expert, medium vs replay), expert behavior vs medium behavior and dedicated data for the task vs exploratory data (plan2explore \citep{plan2explore}). We provide further details on the datasets and histograms of the return distribution in \Cref{sec:datasets}.
In all our experiments the action-conditioned dataset $\mathcal{D}_{act}$ accounts for $5\%$ of the total dataset $\mathcal{D}$.


\label{sec:results_state}

We compare LAWM to several baselines. TD3+BC \citep{TD3BC} is a model-free offline RL approaches based on the idea of adding a behavior cloning penalty to an online RL method \citep{td3} and is trained only on action-labeled data $D_{act}$. We also include an oracle version trained on additional action annotations for the otherwise action-free data $D_{\varnothing}$, providing an approximate performance upper bound. We further evaluate an IDM-assisted variant, where an inverse dynamics model (IDM) trained on $D_{act}$ predicts missing actions in $D_{\varnothing}$ before training TD3+BC on the combined data. Finally, we compare to C-LAP \citep{c-lap}, a model-based approach using the same policy-learning objective as LAWM but an action-conditioned world model, along with its oracle version that also has access to actions in the action-free data.

We report results in \Cref{tab:results_state}. When comparing C-LAP to TD3+BC, we see that the model-based approach achieves stronger results than the model-free baseline especially on datasets with a diverse distribution (medium-replay and plan2explore), but it can fall behind on datasets with a narrow distribution (medium and expert). TD3+BC performs best primarily on expert datasets, which we attribute to the behavior-cloning term in its policy updates as it is very efficient for narrow behavior distributions. However, its effectiveness is constrained by the size of the action-conditioned dataset $D_{act}$, which is evident when comparing it to TD3+BC (Oracle). Adding an inverse dynamics model to label the action-free part of the data (IDM-TD3+BC) improves performance; however, its improvement remains limited and it is not as good as the oracle version. Finally, LAWM clearly benefits from learning jointly from action-free and action-conditioned data: its results indicate that a latent-action world model is superior to training only on action-conditioned trajectories. Moreover, because LAWM also uses an inverse dynamics model to infer latent actions when action labels are absent, these gains suggest that explicitly modeling the system dynamics is important for using action-free data effectively.

\begin{table}[ht]
\centering
\small
\scalebox{0.77}{
\begin{tabular}{cl
                S[table-format = 3.1(2.1)]
                S[table-format = 3.1(2.1)]
                |S[table-format = 3.1(2.1)]
                S[table-format = 3.1(2.1)]
                S[table-format = 3.1(2.1)]
                S[table-format = 3.1(2.1)]}
\toprule
\multicolumn{1}{c}{Environment} & Dataset & {TD3+BC (Oracle)} & {C-LAP (Oracle)} & {TD3+BC} & {C-LAP} & {IDM-TD3+BC} & {LAWM} \\
\cmidrule(lr){3-4}\cmidrule(lr){5-8}
 & & \multicolumn{2}{c|}{\textit{100\% action-conditioned data}} & \multicolumn{4}{c}{\textit{5\% action-conditioned data}} \\
\midrule

\multirow{4}{*}{cheetah-run} 
& expert         & 83.4 \pm 2.4 & 79.6 \pm 2.0 & 52.9 \pm 6.8 & 69.1 \pm 9.4 & \bfseries 85.6 \pm 1.8 & 52.4 \pm 23.5 \\
& medium         & 62.6 \pm 2.1 & 57.8 \pm 3.0 & 32.4 \pm 4.1 & 47.9 \pm 2.5 & \bfseries 61.3 \pm 2.0 & 54.6 \pm 8.7 \\
& medium-replay  & 53.9 \pm 1.8 & 65.0 \pm 8.0 & 14.0 \pm 1.6 & 34.6 \pm 3.4 & 46.6 \pm 19.2 & \bfseries 68.4 \pm 1.4 \\
& plan2explore   &  8.6 \pm 0.3 & 28.5 \pm 6.2 &  4.0 \pm 0.6 & 12.6 \pm 1.2 &  4.8 \pm 4.2 & \bfseries 26.5 \pm 4.6 \\
\midrule

\multirow{4}{*}{hopper-stand} 
& expert         & 72.1 \pm 2.7 & 40.0 \pm 14.6 & 46.5 \pm 5.1 & 13.6 \pm 9.0 & \bfseries 56.5 \pm 8.4 & 37.0 \pm 17.0 \\
& medium         & 45.8 \pm 1.9 & 59.5 \pm 24.9 & 39.6 \pm 12.1 & 60.0 \pm 12.1 & 41.4 \pm 11.5 & \bfseries 65.0 \pm 16.6 \\
& medium-replay  & 15.3 \pm 3.3 & 63.1 \pm 6.5 & 10.2 \pm 3.4 & 19.0 \pm 5.5 & 13.7 \pm 6.9 & \bfseries 46.9 \pm 14.1 \\
& plan2explore   &  2.3 \pm 0.5 & 51.4 \pm 4.8 &  0.8 \pm 2.0 & 27.3 \pm 1.2 &  2.6 \pm 1.2 & \bfseries 54.1 \pm 5.0 \\
\midrule

\multirow{4}{*}{walker-walk} 
& expert         & 92.8 \pm 9.7 & 93.1 \pm 0.5 & 89.6 \pm 3.5 & 88.0 \pm 4.2 & 91.0 \pm 2.2 & \bfseries 94.3 \pm 0.6 \\
& medium         & 71.5 \pm 4.7 & 71.1 \pm 1.0 & 64.7 \pm 1.2 & 68.7 \pm 4.3 & 64.3 \pm 5.5 & \bfseries 91.4 \pm 1.1 \\
& medium-replay  & 63.9 \pm 2.9 & 81.2 \pm 4.6 & 35.9 \pm 2.4 & 75.7 \pm 6.1 & 52.3 \pm 6.3 & \bfseries 75.9 \pm 23.7 \\
& plan2explore   & 18.5 \pm 0.9 & 69.2 \pm 5.4 & 11.8 \pm 7.8 & 34.1 \pm 3.3 &  6.0 \pm 1.9 & \bfseries 81.9 \pm 3.4 \\
\midrule

Average & & 49.2 & 63.3 & 33.5 & 45.9 & 43.8 & \bfseries 62.4 \\
\bottomrule
\end{tabular}
}
\caption{Results on using full-state information on the DeepMind Control Suite. Oracle methods use 100\% of the action dataset, while all other methods use only 5\%. We report mean normalized return $\pm$ standard deviation across multiple seeds and highlight the best performances with bold numbers, excluding Oracle baselines.}
\label{tab:results_state}
\end{table}

\section{Discussion}

In this paper, we address the challenge of using large, action-free datasets with limited action-labeled data for offline reinforcement learning. 
We introduce Latent Action World Models (LAWM), a novel class of world models that learns a latent action space, enabling it to effectively train on these heterogeneous data sources. 
Our experiments on the DeepMind Control Suite demonstrate that LAWM significantly outperforms state-of-the-art baselines.
Notably, it achieves this strong performance while using an order of magnitude less action-conditioned data.
Despite these advances, training a unified inverse dynamics model across diverse behavioral datasets remains challenging. Integrating diverse behaviors into a single latent action representation is a key direction for future work toward more generalizable world models.

\bibliography{iclr2026_conference}
\bibliographystyle{iclr2026_conference}

\newpage
\appendix

\section{Graphical Model}
We provide the graphical model for action-conditioned and the action-free variant of a latent action world model.
\begin{figure}[!h]
\begin{subfigure}[b]{0.5\columnwidth}
  \centering
  \resizebox {0.9\columnwidth}{!} {
      \begin{tikzpicture}
    
      \node[obs]                           (o1) {$o_t$};
      \node[obs, right=2.5cm of o1]              (o2) {$o_{t+1}$};
      \node[latent, right=2.5cm of o2]              (o3) {$o_{t+2}$};
    
      \node[latent, below=2cm of o1]              (s1) {$s_t$};
      \node[latent, below=2cm of o2]              (s2) {$s_{t+1}$};
      \node[latent, below=2cm of o3]              (s3) {$s_{t+2}$};

      \node[latent, right=0.5cm of s1, yshift=1cm]              (u1) {$u_t$};
      \node[latent, right=0.5cm of s2, yshift=1cm]              (u2) {$u_{t+1}$};
              
      \node[obs, above=1cm of u1]              (a1) {$a_{t}$};
      \node[latent, above=1cm of u2]              (a2) {$a_{t+1}$};

      \edge {s1} {o1} ;
      \edge {s2} {o2} ;
      \edge {s3} {o3} ;

      \edge {s1} {s2} ;
      \edge {s2} {s3} ;
      


      \edge {a1} {s2} ;
      \edge {a2} {s3} ;
    
      \edge {u1} {a1} ;
      \edge {u2} {a2} ;
    
      \path (s1) edge[->] (u1) ;
      \path (s1) edge[ ->] (a1) ;
    
      \path (s2) edge[->] (u2) ;
      \path (s2) edge[ ->] (a2) ;
    

      \path (a1) edge[bend left, dashed, ->] (u1) ;
      \path (o1) edge[bend right, dashed, ->] (s1) ;

      \path (o2) edge[bend right=15, dashed, ->] (s2) ;


      \path (s1) edge[bend right, dashed, ->] (u1) ;

      \path (a1) edge[bend left=15, dashed, ->] (s2) ;

      \path (s1) edge[bend right, dashed, ->] (s2) ;
      
    \end{tikzpicture}}

    \subcaption{Action-conditioned \label{fig:graphical_model_action_conditioned}}
\end{subfigure}
\begin{subfigure}[b]{0.5\columnwidth}
  \centering
  \resizebox {0.9\columnwidth}{!} {
      \begin{tikzpicture}
    
      \node[obs]                           (o1) {$o_t$};
      \node[obs, right=2.5cm of o1]              (o2) {$o_{t+1}$};
      \node[latent, right=2.5cm of o2]              (o3) {$o_{t+2}$};
    
      \node[latent, below=2cm of o1]              (s1) {$s_t$};
      \node[latent, below=2cm of o2]              (s2) {$s_{t+1}$};
      \node[latent, below=2cm of o3]              (s3) {$s_{t+2}$};

      \node[latent, right=0.5cm of s1, yshift=1cm]              (u1) {$u_t$};
      \node[latent, right=0.5cm of s2, yshift=1cm]              (u2) {$u_{t+1}$};
              
      \node[latent, above=1cm of u1]              (a1) {$a_{t}$};
      \node[latent, above=1cm of u2]              (a2) {$a_{t+1}$};

      \edge {s1} {o1} ;
      \edge {s2} {o2} ;
      \edge {s3} {o3} ;

      \edge {s1} {s2} ;
      \edge {s2} {s3} ;
      

      \edge {a1} {s2} ;
      \edge {a2} {s3} ;
    
      \edge {u1} {a1} ;
      \edge {u2} {a2} ;
    
      \path (s1) edge[->] (u1) ;
      \path (s1) edge[ ->] (a1) ;
    
      \path (s2) edge[->] (u2) ;
      \path (s2) edge[ ->] (a2) ;
    

      \path (u1) edge[bend right, dashed, ->] (a1) ;
      \path (s1) edge[bend left=15, dashed, ->] (a1) ;
      \path (o1) edge[bend right, dashed, ->] (s1) ;

      \path (o2) edge[bend right=15, dashed, ->] (s2) ;


      \path (s1) edge[bend right, dashed, ->] (u1) ;

      \path (a1) edge[bend left=15, dashed, ->] (s2) ;

      \path (s1) edge[bend right, dashed, ->] (s2) ;

      \path (o2) edge[dashed, ->] (u1) ;

    \end{tikzpicture}}
    \subcaption{Action-free \label{fig:graphical_model_action_free}}
  \end{subfigure}
\caption{We study two settings of latent action world models (LAWM): (a) action-conditioned (actions observed) and (b) action-free. Both models share the same generative model, while the latent action inference model is different. The action-conditioned model infers a latent action $u_t$ given an action $a_t$ from the dataset, while the action-free model needs to rely on $o_{t+1}$ as a weak supervisory signal to infer $u_t$. Solid edges define the generative process and dashed lines the inference process.}
  \label{fig:graphical_model}
\end{figure}
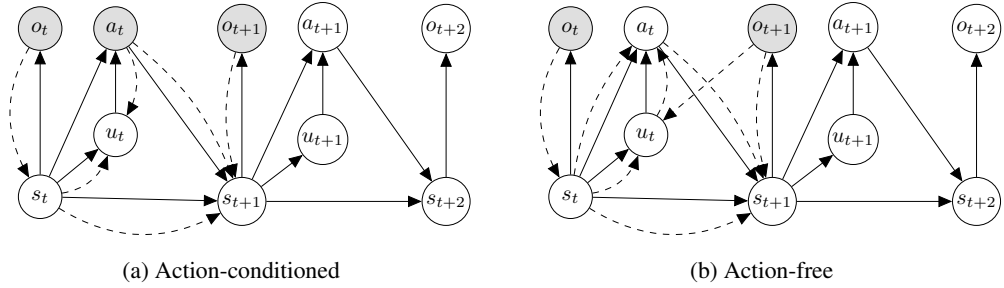

\section{Training objective}
\label{sec:training_objective}
As directly maximizing the marginal likelihood is intractable, we maximize the evidence lower
bound (ELBO) on the log-likelihood. We provide the ELBO for the action-conditioned 

\begin{align}
\begin{split}
 \log p(o_{1:T}, a_{1:T}) &\geq \sum_{t=1}^{T}  \bigl[ \mathbb{E}_{s_t, u_t \sim q_{\phi}} [\underbrace{\log p_{\theta}(o_t | s_t)}_{\text{observation reconstruction}} + \underbrace{\log p_{\theta}(a_t | s_t, u_t)}_{\text{action reconstruction}}] \\&- \underbrace{\displaystyle \KL(q_{\phi}(s_{t} | s_{t-1}, a_{t-1}, o_{t})  \Vert p_{\theta}(s_{t} | s_{t-1}, a_{t-1}))}_{\text{observation consistency}} \\ &- \underbrace{\displaystyle \KL (q_{\phi}(u_{t}| s_{t}, a_t)  \Vert p_{\theta}(u_{t} | s_{t}))}_{\text{action consistency}} \bigr] =: L_{\text{action-conditioned}}
\end{split}
\label{eq:elbo}
\end{align}

and for the action-free setting
\begin{align}
\begin{split}
 \log p(o_{1:T}) &\geq \sum_{t=1}^{T}  \mathbb{E}_{s_t, u_t \sim q_{\phi}} [\underbrace{\log p_{\theta}(o_t | s_t)}_{\text{observation reconstruction}}] \\&- \underbrace{\displaystyle \KL(q_{\phi}(s_{t} | s_{t-1}, a_{t-1}, o_{t})  \Vert p_{\theta}(s_{t} | s_{t-1}, a_{t-1}))}_{\text{observation consistency}} \\ &- \underbrace{\displaystyle \KL(q_{\phi}(u_{t}| s_{t}, o_{t+1})  \Vert p_{\theta}(u_{t}| s_{t}))}_{\text{action consistency}} \\
 &- \underbrace{\displaystyle \KL(q_{\phi}(a_{t}| s_{t}, u_{t})  \Vert p_{\theta}(a_{t}| s_{t}, u_t))}_{\text{action decoder consistency}}\bigr] =: L_{\text{action-free}}.
\end{split}
\label{eq:elbo_af}
\end{align}
We share parameters between $q_{\phi}(a_t | s_t, u_t)$ and $p_{\phi}(a_t | s_t, u_t)$. Therefore, the KL for action decoder consistency can be neglected. Overall, this formulation preserves the training objective of an action-conditioned latent action world model and unifies supervision from mixed datasets with and without actions.

\paragraph{Unified training objective}
Besides settings where only observation trajectories $\mathcal{D}_{\varnothing}=\{(o_{1:T}), ...\}$ or joint observation–action trajectories $\mathcal{D}_{act}=\{(o_{1:T}, a_{1:T}), ...\}$ are available, datasets may also contain a mix of both $\mathcal{D} = \mathcal{D}_{act} \cap \mathcal{D}_{\varnothing}$. To use such a dataset, we define two inference paths: the latent action posterior for action-conditioned data and the action-free latent action posterior for action-free data. During training, the objective is selected depending on whether actions are present in a given trajectory
\begin{align}
L_{\text{model}}(\tau)
&= m(\tau)\,L_{\text{action-conditioned}}(\tau)
  + \bigl(1 - m(\tau)\bigr)\,L_{\text{action-free}}(\tau),\\
\text{where}\quad
m(\tau)
&:= \mathbf{1}\!\left\{\,\forall t \in \{1,\dots,T(\tau)\}:\ a_t \text{ is observed}\,\right\}
\in \{0,1\}.
\end{align}
To use the world model for control and learn a policy using imagined trajectories as in \citep{dreamerv1, LearningToFly}, we learn a reward model that predicts the reward, i.e.
\begin{equation}
    L_{\text{reward}} = \sum_{t=1}^{T} \mathbb{E}_{s_t, u_t \sim q_{\phi}} [\log(p_{\theta}(r_t | s_t))],
    \label{eq:reward_loss}
\end{equation}
leading to the overall objective $ L = L_{\text{model}} + L_{\text{reward}}$.

\section{Implementation Details}
\label{sec:implementation_details}
We follow the implementation of \citet{c-lap}, extending it to the action-free setting. Most hyperparameters for C-LAP and LAWM are shared and listed in \Cref{tab:lawm_hyperparameters}. For TD3+BC, we adopt the hyperparameters proposed in the original implementation \citep{TD3BC}. The IDM for the IDM-TD3+BC baseline is implemented using the hyper-parameters in \Cref{tab:idm_hyperparameters}.

\begin{table}[htb]
\centering
\footnotesize
\setlength{\tabcolsep}{6pt}
\renewcommand{\arraystretch}{1.15}
\begin{tabular}{@{}p{0.40\columnwidth}p{0.45\columnwidth}@{}}
\toprule
\textbf{Model} & \\ 
Stochastic latent state size              & \(64\) \\
Deterministic latent state size           & \(512\) \\
Observation embedding size                & \(512\) \\
Latent action size                        & \(12\) \\
Hidden units                              & \(640\) \\
Hidden activation                         & Swish \citep{swish}  \\
Normalization                             & LayerNorm \citep{ba2016layernormalization}  \\
Free-nats                                 & 1.0 (medium-replay and plan2explore datasets) \\
Encoders and decoders                     & units: \(512\), layers: \(2\)\\
Latent action prior (C-LAP)               & units: \(512\), layers: \(2\)\\
Latent action prior (LAWM)                & Standard normal distribution\\
IDM (LAWM)                                & units: \(512\), layers: \(3\)\\
Learning rate                             & \(3\times10^{-4}\) \\
Batch size                                & \(64\) \\
Window length                             & \(50\) \\

\midrule
\textbf{Agent} & \\
Hidden activation                         & Swish  \\
Normalization                             & LayerNorm  \\
Policy                                    & units: \(256\), layers: \(3\), TanhGaussian distribution \\
Value                                     & units: \(256\), layers: \(3\), number of networks \(2\) \\
Learning rate                             & \(8\times10^{-5}\) \\
Logprob/entropy regulariser               & \(0.01\) \\
Dream rollout length                      & \(5\) \\
Discount factor                           & \(0.99\)\\
Return Estimation                         & Lambda-return \citep{Sutton1998} \\
\bottomrule
\end{tabular}
\caption{LAWM and C-LAP hyper-parameters.}
\label{tab:lawm_hyperparameters}
\end{table}

\begin{table}[htb]
\centering
\footnotesize
\setlength{\tabcolsep}{6pt}
\renewcommand{\arraystretch}{1.15}
\begin{tabular}{@{}p{0.40\columnwidth}p{0.45\columnwidth}@{}}
\toprule
\textbf{Model} & \\ 
MLP                                       & units: \(1024\), layers: \(3\)\\
Hidden activation                         & Swish \citep{swish}  \\
Normalization                             & LayerNorm \citep{ba2016layernormalization}  \\
Dropout                                   & \(0.1\) \citep{dropout} \\
Learning rate                             & \(3\times10^{-4}\) \\
Batch size                                & \(1024\) \\
Window length                             & \(5\) \\
Gradient steps                            & \(10^5\) \\
Learning rate                             & \(1\times10^{-4}\) \\

\bottomrule
\end{tabular}
\caption{IDM hyper-parameters.}
\label{tab:idm_hyperparameters}
\end{table}

\section{LLM Usage}
\label{sec:llm}
Large Language Models (LLMs) are used to assist in the writing and polishing of the manuscript. 
Specifically, we use LLMs for sentence rephrasing, grammar checking to improve the readability of the manuscript. 

\section{Datasets}
\label{sec:datasets}
\begin{figure}[htbp]
    \centering
    \includegraphics[width=1.0\columnwidth]{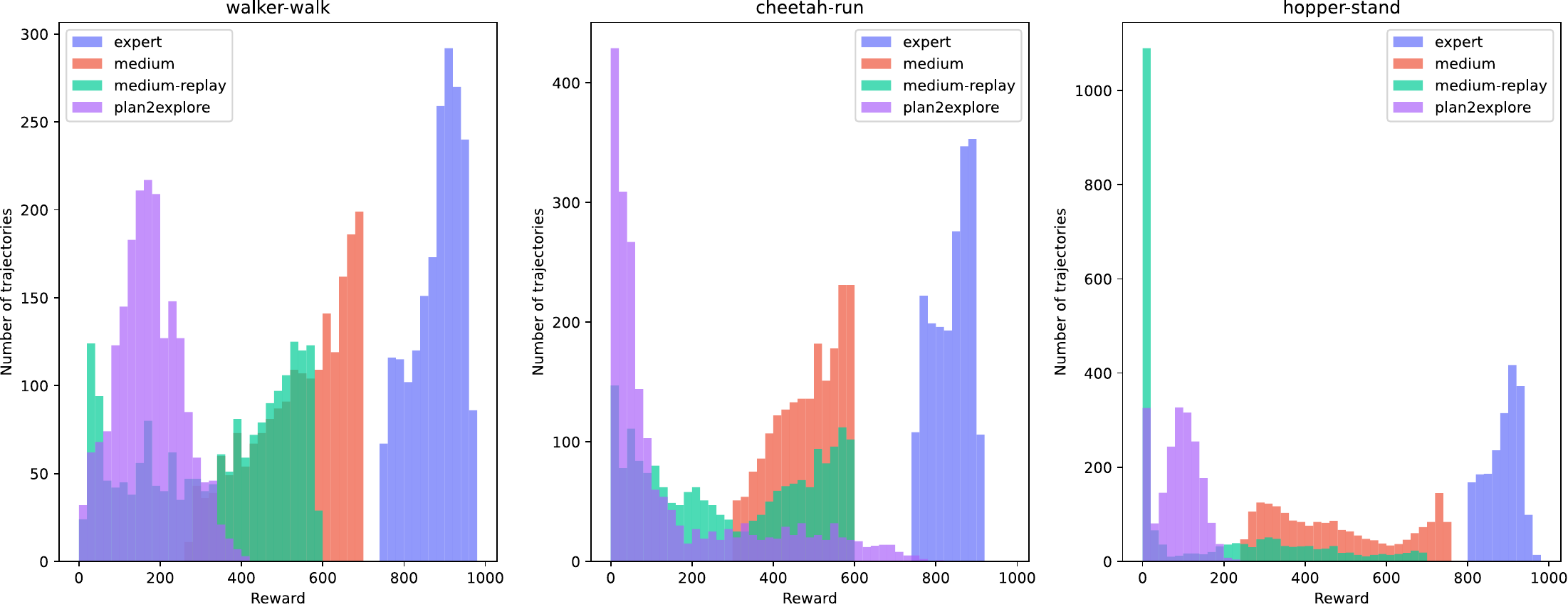}
    \caption{Histograms showing return distributions for the walker-walk, cheetah-run, and hopper-stand environments across four datasets: expert, medium, medium-replay, and the exploratory plan2explore dataset.}
    \label{fig:dataset_hist}
\end{figure}

To create the datasets, we use Dreamer \citep{dreamerv1} to train a policy and collect trajectories (each trajectory has 500 steps). For each environment, we use 5 seeds and gather 3,000 trajectories for hopper-stand and cheetah-run, and 1,500 trajectories for walker-walk per seed while training an agent to expert-level performance. Afterwards, we extract 400 trajectories from each seed with the specified characteristic (medium, medium-replay, expert) and combine the datasets to a total size of 2,000 trajectories. The plan2explore datasets also contain 2,000 trajectories. For plan2explore, we use the dataset from \citep{zhang2025overcomingknowledgebarriersonline}. In \Cref{fig:dataset_hist} we provide histograms of the return distributions and in \Cref{tab:dataset_statistics} the statistics across environments and datasets.

\begin{table*}[htbp]
\centering
\begin{tabular}{
  l
  l
  S[table-format=3.2]  
  S[table-format=3.2]  
  S[table-format=3.2]  
  S[table-format=3.2]  
  S[table-format=3.2]  
  S[table-format=3.2]  
  S[table-format=3.2]  
}
\toprule
{Environment} & {Dataset} & {Mean} & {Std Dev} & {Min} & {P25} & {Median} & {P75} & {Max} \\
\midrule
walker-walk & expert        & 879.45 &  60.44 & 750.17 & 836.05 & 893.35 & 927.55 & 987.15 \\
walker-walk & medium        & 543.44 & 116.24 & 275.70 & 458.50 & 564.06 & 647.34 & 699.98 \\
walker-walk & medium-replay & 333.25 & 182.48 &   4.75 & 167.62 & 372.86 & 501.28 & 584.19 \\
walker-walk & plan2explore  & 173.36 &  79.03 &   4.17 & 119.85 & 169.39 & 227.93 & 418.45 \\
\addlinespace
cheetah-run & expert        & 837.49 &  46.27 & 750.33 & 797.36 & 846.55 & 877.21 & 910.91 \\
cheetah-run & medium        & 483.72 &  80.87 & 301.14 & 421.19 & 495.92 & 555.27 & 599.94 \\
cheetah-run & medium-replay & 297.74 & 200.38 &   2.38 & 100.88 & 286.99 & 494.94 & 599.77 \\
cheetah-run & plan2explore  & 157.74 & 195.37 &   0.00 &  24.18 &  59.74 & 243.13 & 804.35 \\
\addlinespace
hopper-stand & expert        & 887.02 &  41.04 & 800.00 & 856.22 & 894.86 & 919.62 & 983.93 \\
hopper-stand & medium        & 480.67 & 159.36 & 250.57 & 336.79 & 453.52 & 637.16 & 749.98 \\
hopper-stand & medium-replay & 149.22 & 202.90 &   0.00 &   0.00 &   9.21 & 297.64 & 699.79 \\
hopper-stand & plan2explore  &  87.72 &  51.36 &   0.00 &  54.58 &  94.28 & 124.24 & 256.34 \\
\bottomrule
\end{tabular}
\caption{Reward statistics across environments and datasets.}
\label{tab:dataset_statistics}
\end{table*}

\end{document}